\documentclass[lettersize,journal]{IEEEtran}
\usepackage{amsmath,amsfonts}
\usepackage{array}
\usepackage[caption=false,font=normalsize,labelfont=sf,textfont=sf]{subfig}
\usepackage{textcomp}
\usepackage{stfloats}
\usepackage{url}
\usepackage{verbatim}
\usepackage{graphicx}
\usepackage{hyperref}
\usepackage{cleveref}
\usepackage{cite}
\usepackage[T1]{fontenc}
\usepackage{lmodern}
\usepackage{amsmath,amssymb}
\usepackage{bbm}

\usepackage{enumitem}
\usepackage{xurl} 
\usepackage{booktabs}
\usepackage{array}
\usepackage{makecell}
\usepackage[table]{xcolor}
\usepackage{amssymb}
\usepackage{array}
\usepackage{ragged2e}
\usepackage{tabularx}

\newcolumntype{L}[1]{>{\raggedright\arraybackslash}p{#1}}
\newcolumntype{C}[1]{>{\centering\arraybackslash}p{#1}}
\newcolumntype{Y}{>{\RaggedRight\arraybackslash}X}

\usepackage{amsmath,mathtools}
\usepackage{enumitem}

\newcommand{\code}[1]{%
  \ifmmode
    \text{\texttt{\detokenize{#1}}}%
  \else
    \texttt{\detokenize{#1}}%
  \fi
}

\usepackage{listings}
\usepackage[T1]{fontenc}
\usepackage[utf8]{inputenc}
\usepackage{lmodern}
\usepackage{amsmath,amssymb}

\usepackage[utf8]{inputenc}

\usepackage[T1]{fontenc}
\usepackage{lmodern}
\usepackage{amsmath,amssymb}

\usepackage{float}
\usepackage{caption}
\usepackage{fancyvrb}
\usepackage{fvextra} 

\usepackage{algorithm}
\usepackage{algpseudocode}
\usepackage{microtype} 

\makeatletter
\algnewcommand{\StatePar}[1]{%
  \State \parbox[t]{\dimexpr\linewidth-\algorithmicindent\relax}{#1\strut}}
\algnewcommand{\StateParx}[1]{%
  \Statex \parbox[t]{\linewidth}{#1\strut}}
\makeatother

\algrenewcommand\algorithmiccomment[1]{\hfill{\footnotesize$\triangleright$~#1}}

\newcommand{\proc}[1]{\ifmmode\text{\textsc{#1}}\else\textsc{#1}\fi}

\providecommand{\code}[1]{\ifmmode\text{\texttt{#1}}\else\texttt{#1}\fi}

\usepackage{booktabs}
\usepackage{siunitx}
\usepackage{adjustbox} 

\usepackage{booktabs}
\usepackage{tabularx}
\usepackage{makecell}
\usepackage{threeparttable}
\usepackage{pifont}

\begin{document}

\title{Dreaming Across Towns: Semantic Rollout and Town-Adversarial World Models for Zero-Shot Cross-Town Driving in CARLA}

\author{Feeza~Khan~Khanzada,~\IEEEmembership{Student~Member,~IEEE,}
        and Jaerock~Kwon,~\IEEEmembership{Member,~IEEE}%
\thanks{Feeza Khan Khanzada, and Jaerock Kwon are with the Department of Electrical and Computer Engineering, University of Michigan–Dearborn, Dearborn, MI 48128 USA (e-mail: feezakk@umich.edu; jrkwon@umich.edu).}%
\thanks{}%
}

\markboth{}%
{Shell \MakeLowercase{\textit{et al.}}: A Sample Article Using IEEEtran.cls for IEEE Journals}


\maketitle

\begin{abstract}
Driving agents trained in one simulated town often perform poorly in a new town because the road shapes, intersections, and lane layouts can be different. This paper studies how to improve this kind of transfer in the CARLA driving simulator without giving the agent any training data from the test towns. The agent is trained only in Town05 and Town06, then evaluated directly in Town03 and Town04. To focus on road-layout differences, all experiments use the same weather and traffic settings.
We propose a training method that encourages the agent to learn features that are useful across towns rather than features tied to one training town. During training, the agent is asked to predict the high-level visual meaning of future camera views and is also discouraged from relying on cues that reveal which source town the data came from. These extra learning signals are used only during training; at test time, the driving policy uses the same observation and control interface as the baseline agent. In controlled comparisons with matched DreamerV3-style world-model driving agents, the proposed method achieves the highest mean held-out success: 36.6\% on Town03 with a 95\% confidence interval of [30.5, 42.7] and 85.6\% on Town04 with a 95\% confidence interval of [84.0, 87.2], computed across five training seeds. Seed-paired tests against the strongest primary baselines show positive success-rate differences in both held-out towns. Additional experiments show that predicting future visual meaning alone or removing town-specific cues alone is not enough to match the combined method. These results suggest that combining future-scene understanding with reduced reliance on source-town-specific features can improve cross-town driving performance in this CARLA setting.
\end{abstract}


\maketitle

\section{Introduction}

Learned driving agents often degrade when deployed in towns whose road layouts differ from those seen during training. This paper studies zero-shot cross-town transfer for closed-loop driving in CARLA~\cite{dosovitskiy_carla_2017}, an open-source autonomous-driving simulator with configurable sensors, vehicle dynamics, traffic actors, weather, and benchmark towns. CARLA’s town maps contain different road networks and lane layouts, making it useful for controlled studies of cross-town transfer. In our protocol, agents are trained in Town05 and Town06 and evaluated directly in unseen Town03 and Town04 without target-domain adaptation.

We study whether latent representation regularization can improve cross-town closed-loop driving in a Dreamer-style world-model agent. Image augmentation can improve low-level appearance invariance, and domain-adversarial learning can discourage source-domain-specific features, but neither directly constrains how imagined future representations evolve. We therefore ask whether semantic rollout prediction and town-adversarial regularization become more effective when coupled. Semantic prediction shapes imagined latent rollouts, while adversarial training reduces source-town predictability in a semantic latent projection.

The proposed method adds two training-time auxiliary objectives to a DreamerV3-style (a model-based reinforcement-learning agent that learns a recurrent latent world model from high-dimensional observations and trains its actor-critic policy using imagined trajectories in latent space) \cite{hafner_mastering_2024} latent agent. The first predicts frozen OpenCLIP~\cite{cherti_reproducible_2023} image embeddings across logged-action multi-step latent rollouts. The second applies town-adversarial supervision to a semantic projection of the deterministic recurrent state. A causal history-conditioned context feature supports semantic rollout prediction, while the actor and critic retain the standard DreamerV3 control feature. Thus, the semantic and context branches regularize representation learning during training without replacing the control input.

This study focuses on controlled zero-shot cross-town transfer in CARLA. The evaluation isolates structural variation in road layout and lane geometry by using fixed weather, ego-only scenes, and matched Dreamer-family baselines. Within this protocol, we test whether the proposed auxiliary losses improve task completion, collision behavior, lane keeping, and speed behavior relative to comparable DreamerV3-style agents.

\textbf{Scope of this study.} This paper evaluates a fixed-route route-completion protocol, not a full autonomous-driving benchmark. The policy receives no route command, waypoint, target vector, privileged map, traffic actor input, or weather variation. The simulator uses an internal reference path only for reward, termination, and success measurement. Therefore, the claims are limited to whether the proposed auxiliary losses improve a matched Dreamer-style learner under the fixed camera/action interface, source-town training pool, held-out route pools, reward function, checkpoint-selection rule, and evaluation protocol described in \cref{sec:experimental_setup}

The contributions of this paper are threefold.

\begin{itemize}
    \item We propose a semantic-rollout auxiliary objective for Dreamer-style closed-loop driving, in which logged-action latent rollouts are trained to predict frozen OpenCLIP image embeddings over multiple future horizons.

    \item We introduce a semantic-feature town-adversarial regularizer that reduces source-town predictability from a recurrent-state projection while leaving actor-critic control on the standard Dreamer feature.

    \item We provide a controlled CARLA mechanism-isolation study comparing matched Dreamer-family agents under identical source towns, held-out towns, route pools, camera interface, action space, reward, checkpoint-selection rule, and evaluation protocol.

    \item We report seed-paired held-out comparisons and ablations showing that the combined semantic-rollout and source-town-adversarial design improves success over the strongest matched baselines in this protocol.
    
\end{itemize}

The ablation study includes semantic rollout alone, semantic-feature town-adversarial regularization alone, and their combination, allowing us to test whether the two auxiliary signals are complementary under the evaluated protocol.

\section{Related Work}
\label{sec:related_work}

This section reviews work most relevant to the proposed method: learning-based driving, world-model agents, domain generalization, semantic representation learning, and CARLA evaluation protocols.

\subsection{Learning-Based and World-Model Driving}

End-to-end driving policies learn control directly from observations, often with additional conditioning or privileged supervision to reduce ambiguity. Conditional Imitation Learning uses high-level commands to disambiguate maneuvers~\cite{codevilla_end--end_2018}, while Learning by Cheating trains a vision-based student from privileged simulator information~\cite{chen_learning_2019}. Other approaches use compact visual abstractions, reinforcement learning, or hybrid imitation--RL objectives to improve closed-loop behavior~\cite{behl_label_2020,coelho_rlad_2024,chekroun_gri_2022,khanzada_driving_2025}. These methods primarily address policy learning. In contrast, our study focuses on how auxiliary semantic and adversarial losses shape the latent representation of a world-model agent for zero-shot cross-town transfer.

World models learn latent dynamics that support planning, imagination, or model-based value learning from high-dimensional observations. PlaNet and Dreamer showed that recurrent latent dynamics can support planning and policy optimization from pixels~\cite{hafner_learning_2019,hafner_dream_2020}, and DreamerV3 further improved the robustness and scalability of this paradigm~\cite{hafner_mastering_2024}. World-model ideas have also been applied to autonomous driving, including latent neural simulators for CARLA driving, predictive driving models, model-based imitation learning, semantic masked world models, and aligned world models for end-to-end control~\cite{leonardis_think2drive_2025,chen_learning_2021,hu_model-based_2022,gao_enhance_2024,yang_raw2drive_2025,khanzada_indrive_2025}. We test whether semantic rollout supervision and town-adversarial regularization improve transfer when added as auxiliary losses inside a DreamerV3-style closed-loop agent.

\subsection{Domain Shift and Structured Representation Learning}

Robustness to domain shift is a central challenge in learning-based robotics and autonomous driving. Domain randomization improves transfer by varying simulator appearance, lighting, textures, or dynamics during training~\cite{tobin_domain_2017,peng_sim--real_2018,khanzada2024analytical}, while adaptation-based methods infer environment-specific information online~\cite{kumar_rma_2021}. Our setting is different: no target-town data or online adaptation is used, and all methods are evaluated zero-shot in held-out towns.

A related approach is to learn domain-invariant or structured representations. Domain-adversarial training uses gradient reversal to reduce domain predictability while preserving task-relevant information~\cite{ganin_unsupervised_2015}, and invariant or factorized representation methods study how to separate stable factors from nuisance variation~\cite{arjovsky_invariant_2020,bousmalis_domain_2016,higgins_darla_2017,liu_learning_2023}. In our setting, the domain label is the source-town identity, limited to the binary split $\{\text{Town05},\text{Town06}\}$. The proposed town-adversarial loss therefore reduces predictability of the source-town split from a semantic latent projection; it does not directly supervise invariance to held-out towns.

\subsection{Semantic and Vision--Language Supervision}

Semantic abstraction is often used to improve robustness in driving because geometry, lane layout, and scene structure are more transferable than low-level appearance. Vision-language models such as CLIP provide reusable semantic image representations learned from large-scale image--text supervision~\cite{radford_learning_2021}. CLIP-style embeddings have been used in robotics and reinforcement learning as reward signals, representation targets, or high-level semantic priors~\cite{sontakke_roboclip_2023,doroudian_clip-rldrive_2024,chi_learning_2025,dang_clip-motion_2023}. We use frozen OpenCLIP embeddings~\cite{cherti_reproducible_2023} differently: they are not used as rewards or policy inputs. Instead, the world model predicts future OpenCLIP image embeddings along logged-action latent rollouts, encouraging the latent dynamics to preserve semantic information over multiple future horizons.

Unlike CLIP-based reward-shaping or policy-input methods, the proposed method does not expose frozen vision-language embeddings to the policy and does not use them as task rewards. The OpenCLIP embedding is used only as a training-time target for multi-horizon latent rollout prediction. Unlike standard domain-adversarial training, the adversarial loss is applied to a semantic projection that is simultaneously constrained by future semantic prediction. The paper therefore studies the interaction between future semantic rollout supervision and source-town adversarial regularization inside a Dreamer-style world model.

\subsection{CARLA Evaluation Protocols}


Standardized CARLA protocols evaluate full driving stacks under broader combinations of towns, traffic, weather, route commands, and long-horizon scenarios~\cite{delavari_comprehensive_2025}. Our goal is different: we isolate a representation-learning question inside a matched Dreamer-style learner. For this reason, all compared methods share the same simulator, route pools, observation interface, action space, reward, training budget, checkpoint-selection rule, and evaluation metrics. This makes the comparison unsuitable for leaderboard ranking, but appropriate for testing whether the proposed auxiliary losses improve a fixed model-based reinforcement-learning agent.
\section{Methodology}

\subsection{Problem Formulation and Setting}

\begin{figure*}[!t]
    \centering
    \captionsetup{font=footnotesize,skip=2pt}
    \includegraphics[width=\textwidth]{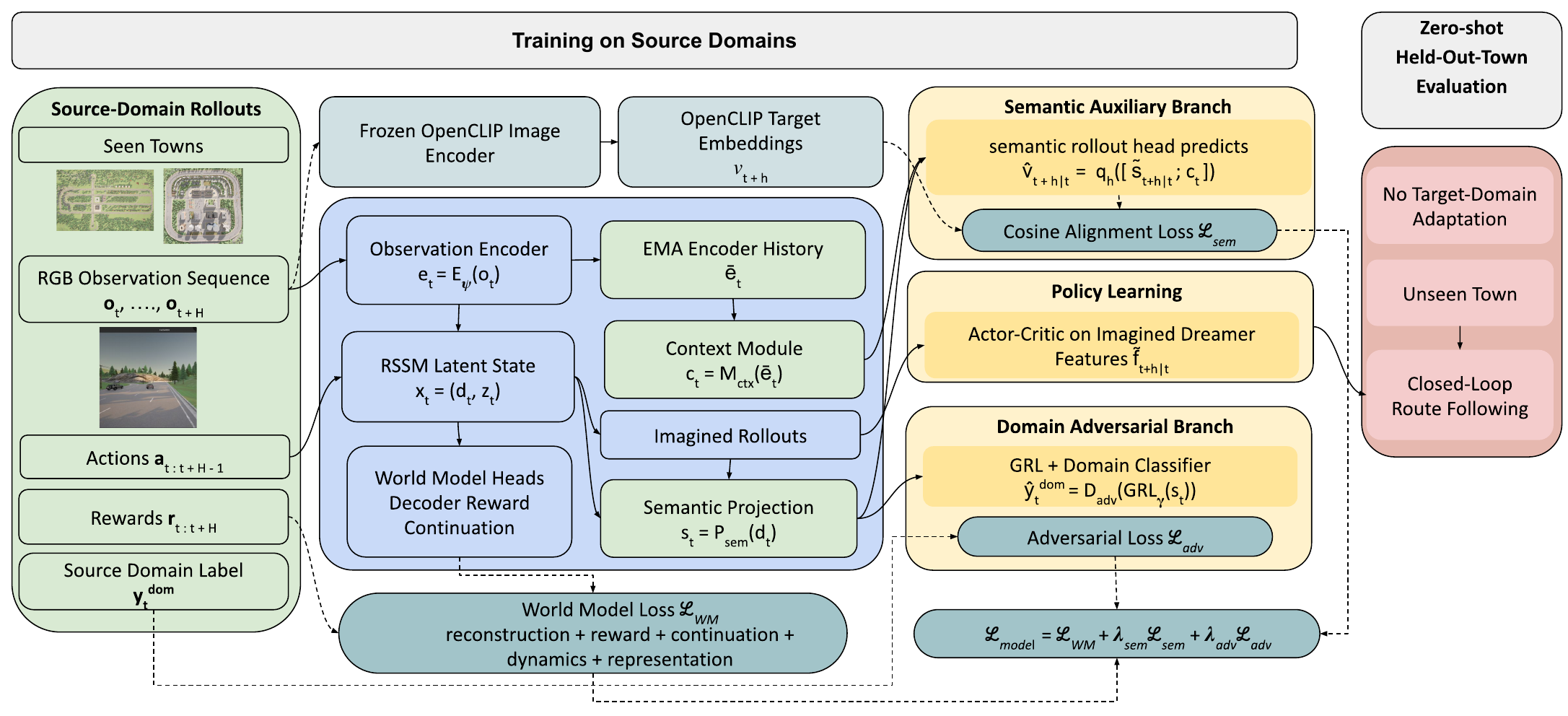}
    \caption{Overview of the proposed framework. The model is trained on source-domain episodes using a Dreamer-style world model, a semantic rollout branch, a policy-learning branch, and a domain-adversarial branch. A frozen OpenCLIP image encoder provides target semantic embeddings for multi-horizon rollout prediction. In the primary model, the actor and critic use the standard Dreamer feature, while the semantic and context branches act as auxiliary training mechanisms. At evaluation time, the learned policy is deployed zero-shot in a held-out town under the same observation-control interface.}
    \label{fig:framework}
\end{figure*}

We formulate each episode as a closed-loop driving task in CARLA. Each episode is initialized from a predefined start pose and terminates upon goal completion or a failure condition such as collision, persistent off-road driving, persistent wrong-lane driving, low-speed timeout, or global timeout. The environment computes progress, reward, and success using an internal reference path, while the learned agent acts from its observation history under a fixed observation-control interface.

We study zero-shot cross-town transfer. Let $D_{\mathrm{src}}$ denote the source towns used for training and $D_{\mathrm{tgt}}$ denote held-out towns used only for evaluation. Training uses trajectories collected in $D_{\mathrm{src}}$. The learned policy is then evaluated directly in $D_{\mathrm{tgt}}$ without target-domain adaptation, fine-tuning, or test-time updates. 

\subsection{Overview of the Proposed Approach}

The proposed method augments a DreamerV3-style latent world model with two training-time auxiliary objectives. The first objective predicts frozen OpenCLIP image embeddings over imagined multi-step rollouts. The second objective applies town-adversarial supervision to a semantic projection of the deterministic RSSM state. A causal history-conditioned context feature summarizes encoder embeddings observed up to time t and conditions the semantic rollout predictor.

\autoref{fig:framework} summarizes the framework. The observation stream is processed by a Dreamer-style world model with recurrent latent dynamics. The standard world-model heads predict observations, rewards, and continuations. In parallel, a semantic rollout branch predicts future OpenCLIP embeddings, and a town-adversarial branch discourages source-town identity from being encoded in the semantic feature. In the primary model, the actor and critic are trained on imagined trajectories using the standard Dreamer feature. The semantic and context branches are, therefore, auxiliary training mechanisms rather than replacements for the control input. At evaluation time, the learned policy is deployed directly in held-out towns without target-domain adaptation, and no OpenCLIP embeddings or town labels are required.

\subsection{Model Architecture} \label{sec:model_architecture}

We use a DreamerV3-style latent world model built from an observation encoder, a recurrent state-space model, and prediction heads for observation reconstruction, reward, and continuation. Let
\[
x_t = (d_t, z_t)
\]
denote the RSSM latent state, where $d_t$ is the deterministic recurrent state and $z_t$ is the stochastic latent state. Given observation $o_t$ and previous action $a_{t-1}$, the encoder produces
\[
e_t = E_\psi(o_t).
\]
The RSSM defines a prior and posterior over latent states as
\[
p_\psi(x_t \mid x_{t-1}, a_{t-1}), 
\qquad 
q_\psi(x_t \mid x_{t-1}, a_{t-1}, e_t).
\]
The standard Dreamer feature $f_t$ is derived from $x_t$ and is used by the actor and critic in the primary model. To construct an auxiliary semantic representation, we map the deterministic recurrent state to a semantic feature,
\[
s_t = P_{\mathrm{sem}}(d_t).
\]
The semantic feature $s_t$ is used by the semantic rollout and domain-adversarial branches. Unless otherwise stated in a dedicated ablation, the actor and critic consume $f_t$; only semantic-policy ablations consume $s_t$.

In parallel, we construct a causal context feature from encoder embeddings rather than from the RSSM state. Let $\bar{e}_t$ denote an episode-reset exponential moving average of encoder embeddings,
\[
\bar{e}_t =
\begin{cases}
e_t, & \text{if } m_t^{\mathrm{first}} = 1,\\
\alpha \bar{e}_{t-1} + (1-\alpha)e_t, & \text{otherwise,}
\end{cases}
\]
where $m_t^{\mathrm{first}} \in \{0,1\}$ indicates whether step $t$ is the first step of a new episode, and $\alpha \in [0,1)$ is a smoothing coefficient. The context feature is
\[
c_t = M_{\mathrm{ctx}}(\bar{e}_t).
\]
Here, $M_{\mathrm{ctx}}$ is a learnable mapping from the encoder-history summary to the context feature. By construction, $c_t$ depends only on observations up to time $t$. In the primary model, $c_t$ conditions the semantic rollout predictor during auxiliary training but is not provided to the actor or critic.

To regularize the latent representation with semantic supervision, we use multi-horizon latent rollouts during world-model training. Starting from the posterior state at time $t$, the RSSM is rolled forward under the logged action sequence $a_{t:t+h-1}$. For horizon $h \in \mathcal{H}$, this produces an imagined latent state
\[
\tilde{x}_{t+h\mid t}
=
\left(\tilde{d}_{t+h\mid t}, \tilde{z}_{t+h\mid t}\right),
\]
where $\mathcal{H} = \{h_1, \ldots, h_m\}$ denotes the set of semantic rollout horizons. The corresponding imagined semantic feature is
\[
\tilde{s}_{t+h\mid t}
=
P_{\mathrm{sem}}\left(\tilde{d}_{t+h\mid t}\right).
\]
A horizon-specific prediction head estimates the frozen OpenCLIP target embedding $v_{t+h}$,
\[
\hat{v}_{t+h\mid t}
=
q_h\left(\left[\tilde{s}_{t+h\mid t}; c_t\right]\right).
\]
This auxiliary branch encourages the imagined latent dynamics to preserve semantic information over multiple future horizons while maintaining a causal dependence on the observation history through $c_t$.

To reduce the predictability of the source-town identity from the semantic representation, we attach a domain classifier to $s_t$ through a gradient-reversal layer,
\[
\hat{y}_t^{\mathrm{dom}}
=
D_{\mathrm{adv}}\left(\mathrm{GRL}_{\gamma}(s_t)\right),
\]
where $y_t^{\mathrm{dom}}$ denotes the source-town label and $\gamma$ controls the adversarial strength. The gradient-reversal operator leaves the forward pass unchanged and multiplies the backward gradient by $-\gamma$. The domain classifier is trained to predict the source town, while the semantic projection receives the reversed gradient and is therefore encouraged to reduce source-town predictability over the binary $\{\text{Town05},\text{Town06}\}$ split. We refer to this objective as town-adversarial regularization rather than town invariance; with only two source towns, the head can at most reduce the predictability of this binary split, and held-out-town invariance is not directly measured by this loss. In the primary model, this adversarial branch is applied to $s_t$, while the actor and critic continue to consume the standard Dreamer feature $f_t$.

\subsection{Learning Objective and Optimization} \label{sec:learning_objective_and_optimization}

The agent is trained from a scalar task reward \(r_t\) provided
by the environment. The reward definition is shared across all methods and is reported in the experimental setup. The learning objective combines the standard Dreamer-style world-model loss with the proposed semantic rollout and town-adversarial auxiliary losses.

\paragraph{World-model objective}
Let \(K\) index the reconstructable observation channels, and let \(o_t^{(k)}\) denote channel \(k\). Let \(f_t\) denote the standard Dreamer feature defined in \cref{sec:model_architecture}. The world model is trained with the standard DreamerV3-style objective
\begin{equation}
\begin{aligned}
\mathcal{L}_{\mathrm{wm}} ={}&
\lambda_{\mathrm{dyn}} \mathcal{L}_{\mathrm{dyn}}
+ \lambda_{\mathrm{rep}} \mathcal{L}_{\mathrm{rep}}
\\
&+ \sum_{k \in K} \lambda_k \, \mathbb{E}\!\left[
-\log p_\phi\!\left(o_t^{(k)} \mid f_t\right)
\right]
\\
&+ \lambda_r \, \mathbb{E}\!\left[
-\log p_\phi(r_t \mid f_t)
\right]
\\
&+ \lambda_\kappa \, \mathbb{E}\!\left[
-\log p_\phi(\kappa_t \mid f_t)
\right].
\end{aligned}
\end{equation}
where \(\kappa_t\) denotes the continuation target, and $\lambda_{dyn}, \lambda_{rep}, \lambda_k, \lambda_r$, and $\lambda_\kappa$ are fixed loss weights. Unless otherwise stated, the dynamics and representation terms follow the standard DreamerV3 objective.

\paragraph{Semantic rollout loss}
Let $H=\{h_1, ..., h_m\}$ denote the set of semantic rollout horizons. For rollout horizons \(h \in \mathcal{H}\), the semantic auxiliary
loss is
\begin{equation}
\mathcal{L}_{\mathrm{sem}}
=
\frac{1}{|\mathcal{H}|}
\sum_{h \in \mathcal{H}}
\mathbb{E}\!\left[
m^{\mathrm{epi}}_{t,h}
\bigl(1 - \cos(\hat v_{t+h \mid t}, v_{t+h})\bigr)
\right].
\end{equation}

Here, $v_{t+h}$ is the frozen OpenCLIP target embedding computed from the future observation, $\hat{v}_{t+h\mid t}$ is the horizon-$h$ prediction defined in \cref{sec:model_architecture}, and $m_{t,h}^{\mathrm{epi}}$ masks terms for which the rollout crosses an episode boundary. Gradients from $\mathcal{L}_{\mathrm{sem}}$ update the RSSM parameters through the imagined rollout, the semantic projection $P_{\mathrm{sem}}$, the context module $M_{\mathrm{ctx}}$, and the prediction head $q_h$. The frozen OpenCLIP encoder is not updated.

\paragraph{Domain-adversarial loss}
Let \(m_t^{\mathrm{dom}} \in \{0,1\}\) indicate whether a source-domain
label is available for sample \(t\). The adversarial domain loss is
\begin{equation}
\mathcal{L}_{\mathrm{adv}}
=
\mathbb{E}\!\left[
m_t^{\mathrm{dom}} \,
\mathrm{CE}\!\left(
D_{\mathrm{adv}}\!\left(\mathrm{GRL}_\gamma(s_t)\right),
y_t^{\mathrm{dom}}
\right)
\right].
\end{equation}

Here, $\mathrm{CE}(\cdot,\cdot)$ denotes cross-entropy, and $\mathrm{GRL}_{\gamma}$ is a gradient-reversal layer with strength $\gamma$. During source-town training, $m_t^{\mathrm{dom}} = 1$ only for samples with an available source-town label $y_t^{\mathrm{dom}}$. No target-town samples are used in this loss.

\paragraph{GRL schedule}

The gradient-reversal coefficient follows
\[
\gamma(p)
=
\gamma_{\max}
\left(
\frac{2}{1+\exp(-10p)} - 1
\right),
\]
where $p \in [0,1]$ is the normalized training progress.

\paragraph{Model objective}
The primary model minimizes
\begin{equation} \label{eq:model_loss}
\mathcal{L}_{\mathrm{model}}
=
\mathcal{L}_{\mathrm{wm}}
+
\lambda_{\mathrm{sem}} \mathcal{L}_{\mathrm{sem}}
+
\lambda_{\mathrm{adv}} \mathcal{L}_{\mathrm{adv}}.
\end{equation}

The shared simulator, training, evaluation, optimization, and auxiliary-loss settings are summarized in \autoref{tab:protocol}.

This objective defines the proposed method. Ablation variants remove selected auxiliary terms or change the actor and critic input, as described in the experimental setup. In particular, $\mathcal{L}_{\mathrm{adv}}$ updates the domain classifier and, through gradient reversal, the semantic branch that produces $s_t$.

Variant-specific losses that are not part of the proposed method are described with the compared methods in \cref{sec:experimental_setup}. The primary method is defined by $\mathcal{L}_{\mathrm{model}}$ and by using the standard Dreamer feature for actor-critic training. Experimental variants modify this design by removing auxiliary losses or changing the actor and critic input, as summarized in \cref{sec:experimental_setup}.

\paragraph{Actor-critic optimization}
Training alternates between minimizing $\mathcal{L}_{\mathrm{model}}$ on replayed trajectories and updating the actor and critic on imagined latent rollouts. Posterior latent states inferred from replay are used as start states for imagination. From each start state, the RSSM is rolled forward for horizon $H_{\mathrm{img}}$ under actions sampled from the actor, producing imagined Dreamer features $\tilde{f}_t$, semantic features $\tilde{s}_t$, predicted rewards $\hat{r}_t$, and continuation probabilities $\hat{c}_t$.

In the primary model, both actor and critic consume the imagined Dreamer feature $\tilde{f}_t$. The critic is trained on $\lambda$-returns computed from imagined rewards and continuation,
\[
\hat{G}_t^{\lambda}
=
\hat{r}_t
+
\hat{c}_t
\left[
(1-\lambda)V_{\eta}^{-}(\tilde{f}_{t+1})
+
\lambda \hat{G}_{t+1}^{\lambda}
\right],
\]
where $V_{\eta}^{-}$ is a slowly updated target critic and $H_{\mathrm{img}}$ denotes the imagination horizon. The actor objective follows the standard DreamerV3-style formulation with entropy regularization. Only dedicated semantic-policy ablations replace $\tilde{f}_t$ with $\tilde{s}_t$ as the actor and critic input. Overall, optimization follows the standard alternation between world-model updates on replayed sequences and behavior learning on imagined latent trajectories.

\section{Experimental Setup} \label{sec:experimental_setup}

\subsection{Simulator and Cross-Town Protocol}

We conducted all experiments in the CARLA simulator~\cite{dosovitskiy_carla_2017} under synchronous simulation with a fixed control step of $0.05$ s, corresponding to a nominal control frequency of 20 Hz. The study evaluated zero-shot cross-town transfer. Agents are trained only in source towns and evaluated in held-out towns without target-domain adaptation.

\begin{figure}[!t]
    \centering
    \captionsetup{font=footnotesize,skip=2pt}
    \includegraphics[width=\columnwidth]{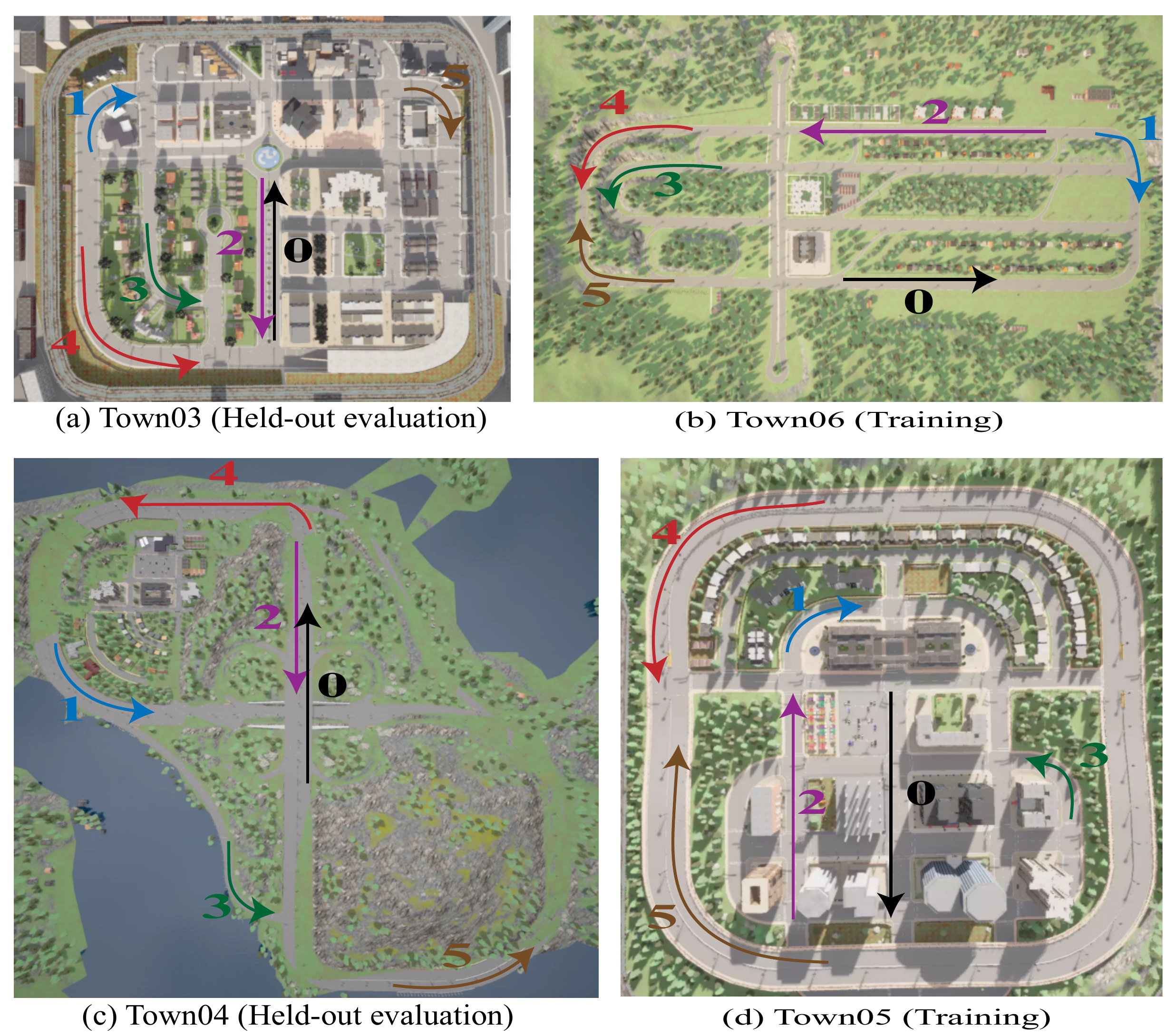}
    \caption{Driving-task maps for the towns used in our experiments. Town05 and Town06 are source training towns; Town03 and Town04 are held-out evaluation towns. Green overlays illustrate the predefined task paths used in the current study.}
    \label{fig:routes}
\end{figure}

The source episode pools contained six predefined driving tasks in Town05 and six in Town06. The held-out evaluation pools contained six predefined driving tasks in Town03 and six in Town04. During training, episodes are sampled uniformly over source towns and then uniformly over tasks within each selected town. Held-out Town03 and Town04 episodes are used only for final evaluation and are never used for training, fine-tuning, or checkpoint selection. \autoref{fig:routes} visualizes the source and held-out task paths used in the study.

\subsection{Observations and Action Space}

The ego vehicle is a Tesla Model 3 controlled directly through CARLA vehicle commands. The primary visual observation is a single front-facing RGB camera mounted on the ego vehicle at approximately $x=1.5\,\mathrm{m}$ and $z=1.6\,\mathrm{m}$, with pitch $-5^\circ$ and field of view $110^\circ$. The camera stream is rendered at $512 \times 512$ resolution and downsampled to $128 \times 128$ before being passed to the learning agent.

The control space is discrete and contains $147$ actions, constructed as the Cartesian product of seven steering bins and twenty-one longitudinal-command bins. The steering set is
\[
[-0.5, -0.25, -0.1, 0.0, 0.1, 0.25, 0.5],
\]
and the longitudinal command set is
\[
[-1.0, -0.9, \ldots, 0.9, 1.0].
\]
Positive longitudinal commands are mapped to throttle, and negative longitudinal commands are mapped to brake magnitude before the resulting control is applied to the simulator. This yields a fixed discrete-action interface shared across all reported methods.

\subsection{Reward and Episode Termination}

The task reward is defined directly in the CARLA environment and is shared across all compared methods. At each control step $t$, the reward is

\begin{equation}
\label{eq:reward}
\begin{aligned}
r_t ={}& 50\Delta \rho_t
- 10 \min(d^{\mathrm{lat}}_t, 1.5)^2
- 20\,\mathbb{1}_{\mathrm{wl}}(t) \\
&- 50 \min(e^{\mathrm{head}}_t, 0.5)^2
-\left(
|v^{\parallel}_t - 5.0|
+ 2|v^{\perp}_t|
+ 2\,\mathbb{1}[v^{\parallel}_t < -0.5]
\right) \\
&+ r^{\mathrm{slow}}_t
- 20\,\mathbb{1}_{\mathrm{off}}(t)
- 5 n^{\mathrm{inv}}_t
- 500 n^{\mathrm{col}}_t
+ 200\,\mathbb{1}_{\mathrm{goal}}(t)
- 0.05 .
\end{aligned}
\end{equation}

where $\rho_t$ denotes progress along the internal reference path, and $\Delta \rho_t = \mathrm{clip}(\rho_t-\rho_{t-1}, -0.5, 0.5)$ is the clipped progress increment. Here, $d^{\mathrm{lat}}_t$ is the lateral distance from the ego vehicle to the internal reference path, $e^{\mathrm{head}}_t$ is the normalized heading error with respect to the local reference tangent, and $v_t^{\parallel}$ and $v_t^{\perp}$ are the longitudinal and lateral velocity components in the path-aligned frame. The binary term $\mathbbm{1}_{\mathrm{wl}}(t)$ indicates a wrong-lane condition, and $\mathbbm{1}_{\mathrm{off}}(t)$ indicates off-road driving. The event counts $n^{\mathrm{inv}}_t$ and $n^{\mathrm{col}}_t$ denote the numbers of new lane-invasion and collision events at the current step. The goal indicator $\mathbbm{1}_{\mathrm{goal}}(t)$ is activated when the distance to the goal falls below $2.0\,\mathrm{m}$.

The low-speed shaping term is

\[
r_t^{\mathrm{slow}}
=
\begin{cases}
0, & v_t \geq 1.0~\mathrm{km/h},\\
-2, & v_t < 1.0~\mathrm{km/h}
\text{ and } N_{\mathrm{slow}} < 10/\Delta t,\\
-50, & v_t < 1.0~\mathrm{km/h}
\text{ and } N_{\mathrm{slow}} \geq 10/\Delta t,
\end{cases}
\]

where $v_t$ is the ego speed in km/h, $N_{\mathrm{slow}}$ is the number of consecutive low-speed steps, and $\Delta t=0.05$ s in our synchronous CARLA setup. It is active when $v_t < 1.0~\mathrm{km/h}$ and inactive otherwise. The low-speed termination condition uses the same threshold, $1.0~\mathrm{km/h}$, for 10 s.

The velocity terms $v_t^{\parallel}$ and $v_t^{\perp}$ are measured in $\mathrm{m/s}$.
The target value $5.0$ in Equation~\ref{eq:reward} therefore corresponds to $5.0~\mathrm{m/s}$.

Episodes terminate upon goal completion, collision, persistent off-road behavior, persistent wrong-lane behavior, persistent low-speed behavior, past-goal overshoot, or timeout. Lane-based termination uses a $1.0\,\mathrm{s}$ grace period after reset. After this grace period, the episode terminates after $1.0\,\mathrm{s}$ of continuous off-road behavior, $3.0\,\mathrm{s}$ of continuous wrong-lane behavior, $10\,\mathrm{s}$ below the low-speed threshold, or $2000$ control steps.

\subsection{Compared Methods and Ablations} \label{sec:methods_ablations}

All compared agents are implemented within the same DreamerV3-style world-model reinforcement-learning framework. They shared the same simulator, task pools, camera configuration, observation interface, action space, reward function, replay pipeline, checkpoint-selection rule, and evaluation protocol. The methods differed only in RGB augmentation, semantic rollout supervision, town-adversarial supervision, and the auxiliary branches enabled during training.

The empirical comparison is designed for controlled mechanism isolation rather than leaderboard-style cross-paradigm ranking. All methods share the same simulator, route pools, observation interface, action space, reward, replay pipeline, checkpoint-selection rule, and evaluation protocol. They differ only in RGB augmentation, semantic rollout supervision, source-town adversarial supervision, and actor-critic feature choice. The study therefore tests whether the proposed auxiliary losses improve a fixed Dreamer-style learner; it does not claim competitiveness with full CARLA driving stacks that use route commands, privileged inputs, planners, imitation learning, multi-sensor perception, or benchmark-specific training pipelines.

\begin{table*}
\caption{Component switches for matched Dreamer-family variants. All methods share the same simulator, route pools, camera configuration, observation interface, world-model backbone, replay pipeline, action space, checkpoint-selection rule, and evaluation protocol.}
\label{tab:variant_summary}

\begin{threeparttable}
\scriptsize
\setlength{\tabcolsep}{2.2pt}
\renewcommand{\arraystretch}{1.08}

\begin{tabularx}{\textwidth}{@{}%
>{\raggedright\arraybackslash}p{2.35cm}
C{0.85cm}
C{0.78cm}
C{0.72cm}
C{0.82cm}
C{0.82cm}
C{0.82cm}
C{0.82cm}
C{0.75cm}
>{\raggedright\arraybackslash}X
@{}}
\toprule
\textbf{Variant} &
\makecell[c]{\textbf{A/C}\\\textbf{input}} &
\makecell[c]{\textbf{Sem.}\\\textbf{proj.}} &
\textbf{Ctx.} &
\makecell[c]{\textbf{Sem.}\\\textbf{roll.}} &
\makecell[c]{\textbf{Base}\\\textbf{adv.}} &
\makecell[c]{\textbf{Sem.}\\\textbf{adv.}} &
\makecell[c]{\textbf{Ctx}\\\textbf{town}} &
\makecell[c]{\textbf{RGB}\\\textbf{aug.}} &
\textbf{Purpose} \\
\midrule

Dreamer-Std
& $f_t$
& No
& No
& No
& No
& No
& No
& No
& Base Dreamer-style world-model RL baseline. \\

Dreamer-Aug
& $f_t$
& No
& No
& No
& No
& No
& No
& Yes
& Tests image-space augmentation without semantic or adversarial losses. \\

Dreamer-DANN
& $f_t$
& No
& No
& No
& Yes
& No
& No
& No
& Tests town-adversarial supervision on the standard Dreamer feature. \\

Sem+Adv-Aux (Ours)
& $f_t$
& Yes
& Yes
& Yes
& No
& Yes
& No
& No
& Main auxiliary design with semantic rollout and semantic-feature town adversary. \\

SemSty-NoLoss
& $f_t$
& Yes
& Yes
& No
& No
& No
& No
& No
& Tests the semantic and context architecture without auxiliary losses. \\

Sem-Rollout-Aux
& $f_t$
& Yes
& Yes
& Yes
& No
& No
& No
& No
& Tests semantic rollout supervision without adversarial regularization. \\

Full-Aux
& $f_t$
& Yes
& Yes
& Yes
& No
& Yes
& Yes
& No
& Tests whether additional context-side town supervision improves the proposed method. \\

Sem-Adv-Aux & $f_t$ & Yes & Yes & No & No & Yes & No & No &
Tests semantic-feature town-adversarial regularization without semantic rollout supervision. \\

Policy-Sem-NoLoss
& $s_t$
& Yes
& Yes
& No
& No
& No
& No
& No
& Tests direct semantic-policy control without auxiliary anchoring. \\

Small-GRL Semantic Policy
& $s_t$
& Yes
& Yes
& Yes
& No
& Yes
& Yes
& No
& Tests direct semantic-policy control with weak adversarial strength. \\

Large-GRL Semantic Policy
& $s_t$
& Yes
& Yes
& Yes
& No
& Yes
& Yes
& No
& Tests direct semantic-policy control with strong adversarial strength. \\

\bottomrule
\end{tabularx}

\begin{tablenotes}[flushleft]
\footnotesize
\item A/C input denotes actor and critic input. $f_t$ is the standard Dreamer feature, and $s_t$ is the semantic feature. Sem. proj. denotes the semantic projection. Ctx. denotes the causal context branch. Sem. roll. denotes semantic rollout supervision. Base adv. denotes town-adversarial supervision on $f_t$. Sem. adv. denotes town-adversarial supervision on $s_t$. Ctx town denotes context-side town supervision. RGB aug. denotes image-space augmentation. The proposed method is Sem+Adv-Aux. Weak and strong GRL settings use different maximum gradient-reversal coefficients, reported in \autoref{tab:protocol}.
\end{tablenotes}

\end{threeparttable}
\end{table*}

\paragraph{Primary baselines}
We used the following primary comparison methods.

\begin{itemize}
    \item \textbf{Dreamer-Std}: the standard DreamerV3-style agent without RGB augmentation, without visual-semantic supervision, and without domain supervision.

    \item \textbf{Dreamer-Aug}: the same base Dreamer agent trained with image-space data augmentation only. In the current configuration, this includes random brightness, contrast, and gamma perturbations, Gaussian blur, and additive image noise.

    \item \textbf{Dreamer-DANN}: the base Dreamer agent with town-domain adversarial
    supervision applied to the standard Dreamer feature $f_t$ through gradient reversal. This baseline uses town labels but does not use the semantic rollout auxiliary branch.

    \item \textbf{Sem+Adv-Aux (Ours):} semantic projection and context branch are enabled together with semantic rollout supervision and semantic-branch town-adversarial supervision. The actor and critic retain the standard Dreamer feature $f_t$, so the semantic/context branch acts as an auxiliary regularizer rather than replacing the base control state. The context-side town-classification loss is disabled.
    
\end{itemize}

\paragraph{Targeted ablations}
To isolate the contribution of each design choice, we also evaluated the following ablations.

\begin{itemize}
    \item \textbf{SemSty-NoLoss:} semantic projection and context branch are enabled, but all semantic and domain auxiliary losses are disabled. This tests whether the architectural change alone affects performance.

    \item \textbf{Sem-Rollout-Aux:} semantic projection and context branch are enabled, and the semantic rollout loss is active, but all domain-related losses are disabled. This isolates the effect of visual-semantic future supervision alone.
    
    \item \textbf{Full-Aux:} extends \textbf{Sem+Adv-Aux (Ours)} by additionally enabling context-side town classification while keeping the actor and critic on the standard Dreamer feature $f_t$. This tests whether the extra context-side domain objective improves over the primary proposed model.

    \item \textbf{Sem-Adv-Aux:} semantic projection and context branch are enabled, semantic rollout supervision is disabled, and semantic-branch town-adversarial supervision is applied to $s_t$. The actor and critic continue to consume the standard Dreamer feature $f_t$. This variant isolates whether adversarial regularization of the semantic branch is useful without semantic rollout supervision.
    
    \item \textbf{Policy-Sem-NoLoss:} semantic projection and context branch are enabled, and the actor/critic consume the semantic feature $s_t$ directly, but all semantic and domain auxiliary losses are disabled. This tests whether simply changing the policy input to the semantic branch is sufficient.
    
    \item \textbf{Small-GRL Semantic Policy:} uses semantic-policy inputs together with semantic rollout supervision, semantic-branch town-adversarial supervision, and context-side town classification, with a weak maximum gradient-reversal strength.
    
    \item \textbf{Large-GRL Semantic Policy:} is identical to the previous variant except for a strong maximum gradient-reversal strength.
\end{itemize}

For town-labeled methods, Town05 and Town06 define the source-domain labels. Held-out-town episodes were not used for optimization.

\subsection{Training Protocol}

All agents were trained exclusively on source-town experience from Town05 and Town06. Held-out Town03 and Town04 data were excluded from all gradient updates, fine-tuning, model selection, and adaptation. For methods that use town supervision, the source-domain labels correspond to the source-town identity.

All compared methods shared the same DreamerV3-style optimization pipeline, including the same world-model updates, replay mechanism, imagined-rollout actor-critic updates, camera preprocessing, discrete action interface, reward definition, and episode termination rules. The methods differed only in the auxiliary components summarized in \autoref{tab:variant_summary}.

\paragraph{Auxiliary branch implementation.}
For semantic-auxiliary variants, the semantic projection \(P_{\mathrm{sem}}\) is implemented as a two-layer MLP applied to the deterministic RSSM state \(d_t\). It uses 512 hidden units, SiLU activations, LayerNorm, and produces a 512-dimensional semantic feature \(s_t\). The context module \(M_{\mathrm{ctx}}\), maps the episode-reset EMA encoder feature \(\bar e_t\) to a 512-dimensional context feature \(c_t\). This context network is also a two-layer MLP with 512 hidden units, SiLU activations, and LayerNorm. The EMA coefficient is \(\alpha=0.9\).

Each horizon-specific semantic prediction head \(q_h\) is a two-layer MLP with 512 hidden units, SiLU activations, and LayerNorm. It takes the semantic feature and context feature,
\((s_{t+h|t},c_t)\), as input and outputs a 512-dimensional prediction of the frozen OpenCLIP image embedding. The prediction and target embedding are L2-normalized inside the
cosine-distance loss. The adversarial classifier \(D_{\mathrm{adv}}\) is implemented as a two-layer MLP with 256 hidden units, SiLU activations, LayerNorm, and a one-hot categorical output over the two source-town labels, Town05 and Town06. In the proposed model, \(D_{\mathrm{adv}}\) is attached to \(s_t\). 

The frozen visual-semantic target is produced using OpenCLIP
\texttt{ViT-B-32} with pretrained weights
\texttt{laion2b\_s34b\_b79k}. RGB camera images are converted to PIL images and passed through the OpenCLIP evaluation preprocessing transform returned by \texttt{open\_clip.create\_model\_and\_transforms}, which performs RGB conversion, bicubic resizing to the ViT-B/32 input resolution, center cropping, tensor conversion, and CLIP
mean/std normalization. The encoded image feature is unit-normalized before being stored in replay. The OpenCLIP encoder is frozen and is not used during evaluation.

The semantic rollout loss \(L_{\mathrm{sem}}\) is included in the world-model objective and its gradients update the observation encoder, RSSM, semantic projection \(P_{\mathrm{sem}}\), context module \(M_{\mathrm{ctx}}\), and horizon-specific prediction heads \(q_h\). The OpenCLIP encoder receives no gradients. The domain-adversarial loss \(L_{\mathrm{adv}}\) updates \(D_{\mathrm{adv}}\) normally and updates \(P_{\mathrm{sem}}\), along with the upstream encoder/RSSM path that produces \(s_t\),
through the gradient-reversal layer. In the primary Sem+Adv-Aux model, the actor and critic use the standard Dreamer feature \(f_t=(d_t,z_t)\); they are not given \(s_t\) or
\(c_t\).

Each method was trained from scratch under five random seeds, $\{2,3,4,5,6\}$. The five-seed training budget was shared across all matched variants for fairness; we therefore report mean and standard deviation across seeds. For methods using semantic rollout supervision, replayed transitions additionally carry frozen OpenCLIP image embeddings. For methods using town supervision, replayed transitions additionally carried the corresponding source-town label and domain mask, where the source-town label space is the binary set $\{\text{Town05},\text{Town06}\}$. Reducing the predictability of this binary label is therefore the operational goal of the town-adversarial loss; held-out-town invariance is not directly supervised and not directly measured during training. Methods that do not use a given auxiliary signal are trained with the corresponding input fields and losses disabled.

Auxiliary loss weights were fixed across seeds. We used $\lambda_{\mathrm{sem}}=0.3$. The town-adversarial loss weight was $\lambda_{\mathrm{adv}}=1.0$ whenever domain-adversarial training is enabled. The context-side town-classification weight is $1.0$ only for Full-Aux and the semantic-policy GRL variants. The gradient-reversal layer followed the DANN schedule and used $\gamma_{\max}=1.0$ for Dreamer-DANN, Sem+Adv-Aux, Sem-Adv-Aux, Full-Aux, and Large-GRL Semantic Policy; $\gamma_{\max}=0.1$ for Small-GRL Semantic Policy; and $\gamma_{\max}=0$ for variants without domain-adversarial training. RGB augmentation was used only in Dreamer-Aug.

For each seed, checkpoints were evaluated every $10\mathrm{K}$ training steps on $20$ closed-loop source-town validation episodes sampled from the Town05 and Town06 route pools. The checkpoint with the highest mean source-validation success was selected for held-out evaluation. Ties were broken by the lower collision rate. No held-out-town data were used for checkpoint selection. \autoref{tab:protocol} summarizes the shared training and evaluation protocol.

\begin{table*}[!t]
\caption{Compact summary of the shared experimental protocol and key hyperparameters. All methods use the same simulator, route pools, observation-control interface, replay pipeline, checkpoint-selection rule, and held-out evaluation protocol. Variant-level component switches are given in \autoref{tab:variant_summary}}
\label{tab:protocol}
\centering
\scriptsize
\setlength{\tabcolsep}{3.2pt}
\renewcommand{\arraystretch}{1.08}
\begin{tabular}{@{}
>{\raggedright\arraybackslash}p{0.17\textwidth}
>{\raggedright\arraybackslash}p{0.31\textwidth}
>{\raggedright\arraybackslash}p{0.17\textwidth}
>{\raggedright\arraybackslash}p{0.31\textwidth}
@{}}
\toprule
\textbf{Setting} & \textbf{Value} & \textbf{Setting} & \textbf{Value} \\
\midrule

CARLA version
& CARLA 0.9.15
& Python / PyTorch
& Python 3.10.16 / PyTorch 2.7.0+cu126 \\

OS / GPU
& Ubuntu 22.04.5 LTS / NVIDIA ADA 6000
& Simulator mode
& Synchronous mode, fixed $\Delta$ t = 0.05 s \\

Source / held-out towns
& Town05, Town06 / Town03, Town04
& Weather / traffic
& ClearNoon; ego vehicle only \\

Routes
& 6 source routes per source town; 6 held-out routes per target town
& Evaluation
& 60 episodes per held-out town per seed; seeds $\{2,3,4,5,6\}$\\

Training budget
& 500K environment steps
& Checkpoint selection
& Every 10K steps using 20 source-town validation episodes; best mean source success, ties by lower collision rate \\

Image and control rate
& $512{\times}512$ rendering, downsampled to $128{\times}128$; 0.05 s control step
& Action space
& 147 discrete actions: 7 steering bins $\times$ 21 longitudinal-command bins \\

Replay
& Capacity $10^5$ transitions; batches of 16 sequences $\times$ 64 steps
& RSSM
& Deterministic size 4096; stochastic size $32{\times}32$ categorical; hidden units 1024; SiLU; LayerNorm \\

Optimization
& Model: Adam, lr $10^{-4}$, $\epsilon=10^{-8}$, clip 1000. Actor/critic: Adam, lr $3{\times}10^{-5}$, $\epsilon=10^{-5}$, clip 100
& Imagination / returns
& $H_{\mathrm{img}}=15$; return $\lambda=0.95$; action-entropy coefficient $3{\times}10^{-4}$ \\

OpenCLIP target
& Frozen OpenCLIP ViT-B/32 image encoder; 512-D unit-normalized target embedding
& Semantic rollout
& Horizons $H=\{1,2,3,4,5\}$; $\lambda_{\mathrm{sem}}=0.3$ for auxiliary rollout variants, $1.0$ for semantic-policy GRL variants, and $0$ otherwise \\

Town labels and domain head
& Two source domains: Town05 and Town06. Domain-head input is $f_t$ for Dreamer-DANN and $s_t$ for semantic-adversarial variants
& Domain-adversarial loss
& $\lambda_{\mathrm{adv}}=1.0$ when enabled; DANN GRL schedule \\

Maximum GRL coefficient
& $\gamma_{\max}=1.0$ for Dreamer-DANN, Sem+Adv-Aux, Sem-Adv-Aux, Full-Aux, and Large-GRL Semantic Policy; $\gamma_{\max}=0.1$ for Small-GRL Semantic Policy; $0$ otherwise
& Context branch
& EMA coefficient $\alpha=0.9$; context-side town loss is $1.0$ for Full-Aux and semantic-policy GRL variants, and $0$ otherwise \\

RGB augmentation
& Dreamer-Aug only: brightness 0.2, contrast 0.2, gamma 0.2, blur probability 0.2, noise std. 0.01
& Held-out-town usage
& No held-out-town data are used for training, fine-tuning, checkpoint selection, or test-time adaptation \\
\bottomrule
\end{tabular}
\end{table*}

\subsection{Evaluation Protocol and Metrics} \label{sec:eval_protocol}

Evaluation was performed in a closed loop on held-out target-town routes without target-domain adaptation, fine-tuning, or test-time updates. Each trained model was evaluated for $60$ closed-loop episodes per held-out town. For each episode, one route was sampled uniformly from the corresponding six-route held-out evaluation pool. Results are reported separately for Town03 and Town04.

An evaluation episode terminates when the agent reaches the goal, collides, remains persistently off-road, remains persistently in the wrong lane, becomes stuck, passes beyond the goal, or reaches the timeout horizon. Success is defined as reaching the goal within a $2\,\mathrm{m}$ radius. The timeout horizon is $2000$ control steps.

We report two groups of metrics. The first group captures task completion and safety:
\begin{itemize}
    \item \textbf{Distance [km over 60 episodes]:} total distance traveled by the ego vehicle accumulated over all $60$ evaluation episodes of a seed.
    \item \textbf{Success [\%]:} percentage of evaluation episodes in which the agent reaches the goal.
    \item \textbf{Collisions/km:} total number of collision events divided by total traveled distance in kilometers.
\end{itemize}

The second group captures lane-keeping and speed behavior, where we used the more sensitive $100\,\mathrm{m}$ scaling to make small differences in lane-discipline visible alongside off-center error, heading error, and speed:
\begin{itemize}
    \item \textbf{Lane invasions/100 m:} total number of lane-invasion events normalized by traveled distance in units of $100\,\mathrm{m}$. This metric is the same quantity as Lane invasions/km up to a fixed factor of ten and is reported here at the $100\,\mathrm{m}$ scale only to keep the lane-keeping table on the same order of magnitude as off-center error and heading error.
    \item \textbf{Off-center error [m]:} mean lateral deviation from the reference route.
    \item \textbf{Heading error [rad]:} mean heading deviation from the local reference direction.
    \item \textbf{Mean speed [km/h]:} episode-averaged vehicle speed, averaged over evaluation episodes.
\end{itemize}

All metrics were first computed for each independently trained seed over the $60$ sampled evaluation episodes of each held-out town. The paper reports the mean and standard deviation across training seeds. We also report 95\% confidence intervals for the primary success metric. Confidence intervals are computed across independently trained seeds using a Student-$t$ interval,
\[
\bar{x} \pm t_{0.975,n-1}\frac{s}{\sqrt{n}},
\]
where $\bar{x}$ is the seed-level mean, $s$ is the seed-level standard deviation, and $n$ is the number of training seeds. The seed, rather than the individual evaluation episode, is treated as the independent
experimental unit. These intervals therefore quantify variability due to training randomness under the
fixed evaluation protocol. For the primary success metric, we additionally retain raw seed-level success counts and perform seed-paired comparisons. Let $y_{i,m,t}$ denote the number of successful episodes out of 60 for seed $i$, method $m$, and held-out town $t$. The corresponding success rate is $p_{i,m,t}=100y_{i,m,t}/60$. For a baseline $b$, the paired difference is
\[
\Delta_{i,b,t}=p_{i,\mathrm{ours},t}-p_{i,b,t}.
\]
We report the mean paired difference, a 95\% Student-$t$ confidence interval, and a two-sided paired $t$-test across training seeds. The trained seed, rather than the individual evaluation episode, is treated as the independent experimental unit. For distance-normalized event rates, event counts, and traveled distance were accumulated over all evaluation episodes of a seed before normalization. This avoids distortions caused by averaging per-episode ratios over episodes with different path lengths.

\subsection{Reproducibility}

To make the custom CARLA protocol reproducible, we provide the route-definition files, source-validation episode lists, held-out evaluation episode lists, training seeds, evaluation seeds, checkpoint-selection logs, and scripts used to compute Tables 3–7. The route files specify the start pose, goal pose, route waypoints, and route identifier for each R0–R5 task in every town. We also provide the CARLA version, Python/PyTorch versions, simulator synchronous-mode settings, sensor configuration, action discretization, and OpenCLIP preprocessing code. No held-out-town data are included in training, fine-tuning, checkpoint selection, or test-time adaptation.
\section{Results}
\label{sec:results}

\begin{table*}[t]
\centering
\captionsetup{font=footnotesize,skip=2pt}
\scriptsize
\setlength{\tabcolsep}{4.2pt}
\renewcommand{\arraystretch}{1.12}
\caption{Town03 held-out evaluation. Metrics are reported as mean $\pm$ standard deviation across five training seeds, with 60 sampled evaluation episodes per seed. Success is the primary metric. Lane invasions are reported per 100 m to avoid duplicating the equivalent lane-invasions/km statistic.}
\label{tab:town3_compact}
\begin{tabular}{@{}lccccccc@{}}
\toprule
\textbf{Method} &
\makecell{\textbf{Success}\\\textbf{[\%]}$\uparrow$} &
\makecell{\textbf{Distance}\\\textbf{[km / 60 eps]}} &
\makecell{\textbf{Coll.}\\\textbf{/km}$\downarrow$} &
\makecell{\textbf{Lane inv.}\\\textbf{/100m}$\downarrow$} &
\makecell{\textbf{Off-center}\\\textbf{[m]}$\downarrow$} &
\makecell{\textbf{Heading err.}\\\textbf{[rad]}$\downarrow$} &
\makecell{\textbf{Mean speed}\\\textbf{[km/h]}} \\
\midrule
Sem+Adv-Aux (Ours)        & $\mathbf{36.6 \pm 4.9}$ & $5.17 \pm 0.28$ & $0.05 \pm 0.09$ & $\mathbf{9.95 \pm 0.94}$ & $\mathbf{0.605 \pm 0.044}$ & $\mathbf{0.0232 \pm 0.0012}$ & $15.99 \pm 0.31$ \\
Dreamer-Aug               & $29.5 \pm 1.5$          & $4.57 \pm 0.16$ & $0.04 \pm 0.09$ & $12.62 \pm 0.58$ & $0.900 \pm 0.022$ & $0.0260 \pm 0.0006$ & $18.75 \pm 0.49$ \\
Dreamer-DANN              & $28.3 \pm 3.9$          & $5.29 \pm 0.35$ & $0.00 \pm 0.00$ & $11.48 \pm 1.10$ & $0.928 \pm 0.053$ & $0.0279 \pm 0.0012$ & $17.23 \pm 0.38$ \\
Dreamer-Std               & $5.3 \pm 2.3$           & $3.97 \pm 0.15$ & $0.02 \pm 0.08$ & $12.08 \pm 0.78$ & $1.942 \pm 0.100$ & $0.0341 \pm 0.0014$ & $23.83 \pm 0.38$ \\
\midrule
Sem-Rollout-Aux           & $12.5 \pm 2.9$          & $5.06 \pm 0.18$ & $0.75 \pm 0.44$ & $11.27 \pm 0.51$ & $1.158 \pm 0.062$ & $0.0252 \pm 0.0010$ & $22.42 \pm 0.31$ \\
Sem-Adv-Aux               & $0.0 \pm 0.0$           & $1.34 \pm 0.04$ & $\mathbf{0.00 \pm 0.00}$ & $22.40 \pm 1.00$ & $0.704 \pm 0.022$ & $0.0289 \pm 0.0013$ & $0.91 \pm 0.01$ \\
SemSty-NoLoss             & $10.5 \pm 5.5$          & $4.11 \pm 0.26$ & $\mathbf{0.00 \pm 0.00}$ & $11.48 \pm 0.69$ & $1.139 \pm 0.055$ & $0.0262 \pm 0.0014$ & $22.05 \pm 0.39$ \\
Full-Aux                  & $8.2 \pm 2.6$           & $4.15 \pm 0.48$ & $0.04 \pm 0.10$ & $16.68 \pm 1.15$ & $1.346 \pm 0.114$ & $0.0319 \pm 0.0017$ & $16.60 \pm 0.60$ \\
\midrule
Policy-Sem-NoLoss         & $0.1 \pm 0.5$           & $3.52 \pm 0.13$ & $0.00 \pm 0.00$ & $14.19 \pm 0.85$ & $1.003 \pm 0.058$ & $0.0372 \pm 0.0040$ & $10.38 \pm 0.77$ \\
Small-GRL Semantic Policy & $3.1 \pm 2.0$           & $4.74 \pm 0.18$ & $0.62 \pm 0.43$ & $13.62 \pm 0.84$ & $1.750 \pm 0.127$ & $0.0365 \pm 0.0025$ & $23.51 \pm 0.59$ \\
Large-GRL Semantic Policy & $0.7 \pm 1.1$           & $4.07 \pm 0.19$ & $0.67 \pm 0.28$ & $14.22 \pm 0.83$ & $1.570 \pm 0.077$ & $0.0334 \pm 0.0014$ & $19.60 \pm 0.62$ \\
\bottomrule
\end{tabular}
\end{table*}
\begin{table*}[t]
\centering
\captionsetup{font=footnotesize,skip=2pt}
\scriptsize
\setlength{\tabcolsep}{4.2pt}
\renewcommand{\arraystretch}{1.12}
\caption{Town04 held-out evaluation. Metrics are reported as mean $\pm$ standard deviation across five training seeds, with 60 sampled evaluation episodes per seed. Success is the primary metric. Lane invasions are reported per 100 m to avoid duplicating the equivalent lane-invasions/km statistic.}
\label{tab:town4_compact}
\begin{tabular}{@{}lccccccc@{}}
\toprule
\textbf{Method} &
\makecell{\textbf{Success}\\\textbf{[\%]}$\uparrow$} &
\makecell{\textbf{Distance}\\\textbf{[km / 60 eps]}} &
\makecell{\textbf{Coll.}\\\textbf{/km}$\downarrow$} &
\makecell{\textbf{Lane inv.}\\\textbf{/100m}$\downarrow$} &
\makecell{\textbf{Off-center}\\\textbf{[m]}$\downarrow$} &
\makecell{\textbf{Heading err.}\\\textbf{[rad]}$\downarrow$} &
\makecell{\textbf{Mean speed}\\\textbf{[km/h]}} \\
\midrule
Sem+Adv-Aux (Ours)        & $\mathbf{85.6 \pm 1.3}$ & $6.62 \pm 0.09$ & $\mathbf{0.00 \pm 0.00}$ & $10.97 \pm 0.75$ & $\mathbf{0.590 \pm 0.037}$ & $0.0201 \pm 0.0016$ & $14.60 \pm 0.73$ \\
Dreamer-Aug               & $80.0 \pm 2.0$          & $6.35 \pm 0.06$ & $\mathbf{0.00 \pm 0.00}$ & $10.57 \pm 1.23$ & $\mathbf{0.514 \pm 0.017}$ & $\mathbf{0.0146 \pm 0.0004}$ & $20.70 \pm 0.12$ \\
Dreamer-DANN              & $82.5 \pm 2.6$          & $6.60 \pm 0.11$ & $\mathbf{0.00 \pm 0.00}$ & $10.83 \pm 0.71$ & $0.683 \pm 0.032$ & $0.0156 \pm 0.0010$ & $17.89 \pm 0.22$ \\
Dreamer-Std               & $33.4 \pm 0.9$          & $5.13 \pm 0.12$ & $\mathbf{0.00 \pm 0.00}$ & $\mathbf{8.78 \pm 0.42}$  & $1.148 \pm 0.026$ & $0.0174 \pm 0.0002$ & $25.50 \pm 0.43$ \\
\midrule
Sem-Rollout-Aux           & $53.4 \pm 2.6$          & $5.30 \pm 0.05$ & $0.11 \pm 0.10$ & $9.78 \pm 0.64$  & $0.820 \pm 0.032$ & $0.0152 \pm 0.0006$ & $20.19 \pm 0.26$ \\
Sem-Adv-Aux               & $0.0 \pm 0.0$           & $1.67 \pm 0.03$ & $0.61 \pm 0.61$ & $28.85 \pm 1.70$ & $0.659 \pm 0.037$ & $0.0268 \pm 0.0017$ & $1.18 \pm 0.05$ \\
SemSty-NoLoss             & $54.7 \pm 2.2$          & $6.19 \pm 0.12$ & $1.55 \pm 0.39$ & $11.49 \pm 0.44$ & $0.968 \pm 0.023$ & $0.0219 \pm 0.0006$ & $27.08 \pm 0.29$ \\
Full-Aux                  & $56.9 \pm 1.8$          & $4.85 \pm 0.13$ & $\mathbf{0.00 \pm 0.00}$ & $13.36 \pm 0.91$ & $0.705 \pm 0.013$ & $0.0183 \pm 0.0005$ & $17.34 \pm 0.11$ \\
\midrule
Policy-Sem-NoLoss         & $38.8 \pm 2.6$          & $4.85 \pm 0.06$ & $\mathbf{0.00 \pm 0.00}$ & $11.55 \pm 0.55$ & $0.950 \pm 0.016$ & $0.0171 \pm 0.0005$ & $18.09 \pm 0.35$ \\
Small-GRL Semantic Policy & $49.1 \pm 2.1$          & $5.22 \pm 0.08$ & $\mathbf{0.00 \pm 0.00}$ & $10.66 \pm 0.58$ & $0.868 \pm 0.024$ & $0.0175 \pm 0.0006$ & $22.96 \pm 0.55$ \\
Large-GRL Semantic Policy & $53.8 \pm 3.4$          & $5.99 \pm 0.14$ & $0.70 \pm 0.21$ & $10.41 \pm 0.46$ & $0.914 \pm 0.031$ & $0.0182 \pm 0.0005$ & $28.32 \pm 0.24$ \\
\bottomrule
\end{tabular}
\end{table*}
\begin{table}[t]
\centering
\caption{95\% confidence intervals for held-out success. Intervals are computed across independently trained seeds using 
$\bar{x} \pm t_{0.975,n-1}s/\sqrt{n}$ with $n=5$. Success is reported in percent.}
\label{tab:success_ci}
\scriptsize
\setlength{\tabcolsep}{2.5pt}
\renewcommand{\arraystretch}{1.05}
\begin{tabular}{lcc}
\hline
\textbf{Method} & \textbf{Town03} & \textbf{Town04} \\
\hline
Sem+Adv-Aux (Ours) & \textbf{36.6 [30.5, 42.7]} & \textbf{85.6 [84.0, 87.2]} \\
Dreamer-Aug & 29.5 [27.6, 31.4] & 80.0 [77.5, 82.5] \\
Dreamer-DANN & 28.3 [23.5, 33.1] & 82.5 [79.3, 85.7] \\
Dreamer-Std & 5.3 [2.4, 8.2] & 33.4 [32.3, 34.5] \\
Sem-Rollout-Aux & 12.5 [8.9, 16.1] & 53.4 [50.2, 56.6] \\
Sem-Adv-Aux & 0.0 [0.0, 0.0] & 0.0 [0.0, 0.0] \\
SemSty-NoLoss & 10.5 [3.7, 17.3] & 54.7 [52.0, 57.4] \\
Full-Aux & 8.2 [5.0, 11.4] & 56.9 [54.7, 59.1] \\
Policy-Sem-NoLoss & 0.1 [0.0, 0.7] & 38.8 [35.6, 42.0] \\
Small-GRL Sem. Policy & 3.1 [0.6, 5.6] & 49.1 [46.5, 51.7] \\
Large-GRL Sem. Policy & 0.7 [0.0, 2.1] & 53.8 [49.6, 58.0] \\
\hline

\end{tabular}
\end{table}
\begin{table}[t]
\centering
\caption{Seed-paired success-rate differences for the primary held-out comparisons. Differences are computed as Sem+Adv-Aux minus the baseline in percentage points. Confidence intervals use a Student-$t$ interval over paired seed-level differences.}
\label{tab:paired_success}
\scriptsize
\setlength{\tabcolsep}{2.5pt}
\renewcommand{\arraystretch}{1.05}
\begin{tabular}{llcccc}
\hline
\textbf{Town} & \textbf{Baseline} & $\bar{\Delta}$ \textbf{[pp]} & \textbf{95\% CI [pp]} & $t(4)$ & $p$ \\
\hline
Town03 & Dreamer-Aug  & 7.1 & [2.6, 11.6] & 4.4  & 0.012 \\
Town03 & Dreamer-DANN & 8.3 & [6.8, 9.8]  & 15.4 & $<0.001$ \\
Town04 & Dreamer-Aug  & 5.6 & [4.5, 6.7]  & 14.1 & $<0.001$ \\
Town04 & Dreamer-DANN & 3.1 & [1.3, 4.9]  & 4.8  & 0.009 \\
\hline
\end{tabular}
\end{table}

We evaluated zero-shot cross-town transfer in a closed loop on held-out Town03 and Town04. No held-out-town data were used for adaptation, fine-tuning, checkpoint selection, or test-time updates. \autoref{tab:town3_compact} and \ref{tab:town4_compact} report task-completion, safety, lane-keeping, and speed metrics. \autoref{fig:success_ft_vs_st} summarizes held-out success rates, \autoref{fig:failure_modes} reports termination outcomes, and \autoref{fig:train_diagnostics_appendix} reports source-training auxiliary diagnostics.

All values in \autoref{tab:town3_compact} and \ref{tab:town4_compact} are reported as mean $\pm$ standard deviation across five independently trained seeds.

\subsection{Main Held-Out-Town Transfer Results}

To assess whether the success gains are consistent across training seeds, \autoref{tab:paired_success} reports seed-paired success-rate differences between Sem+Adv-Aux and the two strongest primary baselines, Dreamer-Aug and Dreamer-DANN. Differences are computed in percentage points as Sem+Adv-Aux minus the corresponding baseline for the same training seed and held-out town. All paired confidence intervals exclude zero. On Town03, Sem+Adv-Aux improves over Dreamer-Aug by 7.1 percentage points and over Dreamer-DANN by 8.3 percentage points on average. On Town04, the corresponding paired mean gains are 5.6 percentage points over Dreamer-Aug and 3.1 percentage points over Dreamer-DANN. The Town04 comparison with Dreamer-DANN is the smallest margin, but the paired interval remains positive. Because the paired tests use five training seeds and multiple baseline comparisons, 
the p-values are reported as descriptive statistics; the primary evidence is the 
direction and magnitude of the seed-paired differences together with the held-out 
closed-loop success rates.

\autoref{tab:town3_compact} and~\ref{tab:town4_compact} report held-out closed-loop performance on Town03 and Town04. Across the matched Dreamer-family variants, Sem+Adv-Aux obtains the highest mean success in both target towns. On Town03, Sem+Adv-Aux reaches 36.6 ± 4.9\% success, with a 95\% confidence interval of [30.5, 42.7]. The strongest baselines are Dreamer-Aug, with 29.5 ± 1.5\% success and a 95\% confidence interval of [27.6, 31.4], and Dreamer-DANN, with 28.3 ± 3.9\% success and a 95\% confidence interval of [23.5, 33.1]. On Town04, Sem+Adv-Aux reaches $85.6 \pm 1.3$\% success, with a 95\% confidence interval of [84.0, 87.2], compared with $80.0 \pm 2.0$\% [77.5, 82.5] for Dreamer-Aug and $82.5 \pm 2.6$\% [79.3, 85.7] for Dreamer-DANN, as reported in \autoref{tab:success_ci}.

The Town03 result shows the larger practical gain over the strongest baselines. The Town04 result has a smaller absolute margin over Dreamer-DANN, but the seed-paired confidence interval remains above zero. We therefore interpret the evidence as supporting a positive seed-paired success-rate differences within the tested five seeds over the strongest primary baselines within this controlled protocol, while noting that the practical Town04 margin over Dreamer-DANN is modest.

\begin{figure*}[!t]
    \centering
    \captionsetup{font=footnotesize,skip=2pt}
    \includegraphics[width=\textwidth]{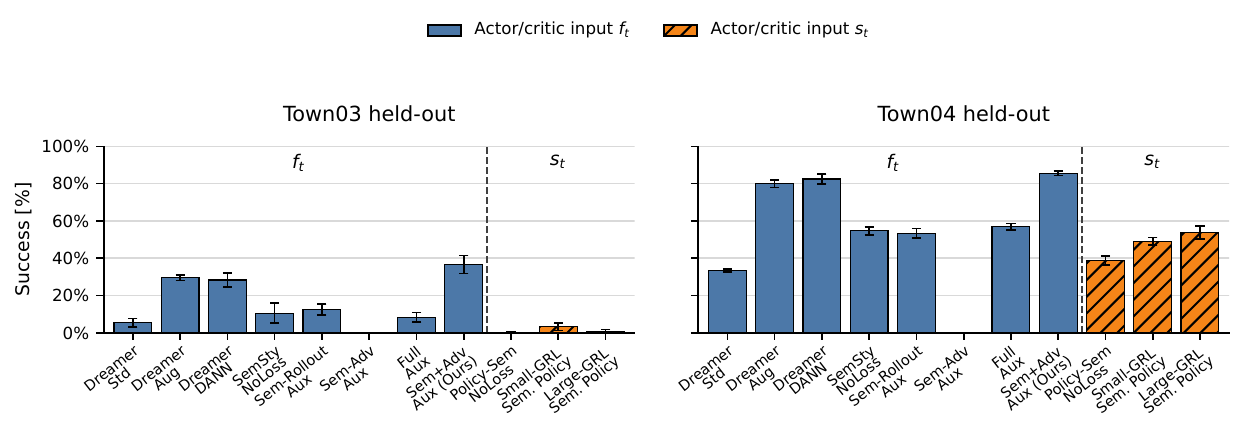}
    \caption{Held-out-town success rates on Town03 and Town04. Bars show mean success across five training seeds; error bars show 95\% confidence intervals computed across seeds using a Student-t interval. Methods are grouped by whether the actor and critic consume the standard Dreamer feature $f_t$ or the semantic feature $s_t$. The grouping is descriptive because variants also differ in auxiliary losses. The strongest results occur when the semantic and context branches are used as auxiliary regularizers while control remains on $f_t$.}
    \label{fig:success_ft_vs_st}
\end{figure*}

\autoref{fig:success_ft_vs_st} visualizes the same pattern. The proposed method ranks first on both Town03 and Town04. Dreamer-Aug and Dreamer-DANN are competitive, especially on Town04, but neither matches the proposed method across both held-out towns. The semantic-policy variants remain weak on Town03, showing that directly routing control through the semantic feature $s_t$ is not sufficient for transfer.

\subsection{Ablation Analysis}

The ablation results indicate that the two auxiliary losses are complementary under the tested hyperparameters. Semantic rollout supervision alone improves over Dreamer-Std but remains below Sem+Adv-Aux. Semantic-feature town-adversarial regularization without semantic rollout collapses, producing near-zero success and very low speed. This suggests that adversarial pressure on the semantic branch is harmful unless the branch is simultaneously anchored by future semantic prediction.

The results show that neither auxiliary mechanism is sufficient by itself. \textbf{Sem-Rollout-Aux}, which uses semantic rollout supervision without adversarial regularization, improves over Dreamer-Std but remains far below the proposed method. It reaches $12.5 \pm 2.9\%$ success on Town03 and $53.4 \pm 2.6\%$ on Town04, compared with $36.6 \pm 4.9\%$ and $85.6 \pm 1.3\%$ for Sem+Adv-Aux. On Town03, Sem-Rollout-Aux also incurs a much higher collision rate than the proposed method, indicating that semantic prediction alone is not a benign partial version of the full model.

The \textbf{Sem-Adv-Aux} ablation was even more diagnostic. It obtained $0.0\%$ success on both held-out towns, traveled only $1.34$ km on Town03 and $1.67$ km on Town04 over $60$ episodes, and had a very low mean speed: approximately $0.91$ km/h on Town03 and $1.18$ km/h on Town04. This indicates a degenerate behavior mode under the tested hyperparameters. The semantic-feature adversarial objective by itself appears to suppress or distort information needed for useful closed-loop control unless it is anchored by semantic rollout prediction.

The comparison between \textbf{Dreamer-DANN} and \textbf{Sem-Adv-Aux} further clarifies the mechanism. Dreamer-DANN applies town-adversarial supervision to the standard Dreamer feature $f_t$ and remains a strong baseline, especially on Town04. Sem-Adv-Aux applies adversarial supervision to $s_t$ without semantic rollout anchoring and collapses. The proposed method applies the adversarial loss to $s_t$ but also constrains $s_t$ through multi-horizon semantic rollout prediction. The resulting performance supports the interpretation that, under the tested hyperparameters, semantic-feature adversarial regularization is most effective when the semantic branch is also anchored by multi-horizon semantic rollout prediction.

Architecture-only and policy-input ablations reinforce this conclusion. \textbf{SemSty-NoLoss}, which adds the semantic and context branches without auxiliary losses, improves over Dreamer-Std but remains well below the proposed method. \textbf{Policy-Sem-NoLoss}, which feeds $s_t$ directly to the actor and critic without auxiliary anchoring, performs poorly, especially on Town03. The small-GRL and large-GRL semantic-policy variants are also weaker than the proposed method. Thus, the semantic branch is most effective as an auxiliary regularizer, not as a direct replacement for the standard Dreamer control feature.

Finally, \textbf{Full-Aux} performs substantially worse than Sem+Adv-Aux despite enabling additional context-side town supervision. This suggests that adding more domain-related supervision is not automatically beneficial. Within the tested family, the best configuration is the simpler one: semantic rollout prediction plus semantic-feature town adversarial regularization, while actor and critic continue to consume $f_t$.

\subsection{Lane-Keeping and Speed Behavior}

\autoref{tab:town3_compact} and~\ref{tab:town4_compact} provide a finer-grained view of lane-keeping and speed behavior. These metrics are important because a method can achieve low collision rates by moving slowly, stopping, or terminating early rather than by driving well.

On Town03, Sem+Adv-Aux has the best overall lane-keeping profile among the successful methods. It records approximately $9.95$ lane invasions per $100$ m, $0.605$ m mean off-center error, and $0.0232$ rad heading error. These values are lower than the corresponding Dreamer-Aug and Dreamer-DANN values in \autoref{tab:town3_compact}. This supports the interpretation that the Town03 success gain is associated with better route adherence rather than merely more aggressive progress.

On Town04, the secondary metrics are more mixed. Dreamer-Aug has slightly lower off-center and heading error, and Sem-Rollout-Aux and Dreamer-Std have lower lane-invasion rates per $100$ m. However, these methods do not match the proposed method's success rate. This is a useful trade-off to state explicitly: lower values on a local lane-keeping metric do not necessarily imply better long-horizon closed-loop task completion.

The speed metrics show that Sem+Adv-Aux is relatively conservative. On Town03, it drives slower than several baselines while still achieving the highest success. On Town04, it is slower than Dreamer-Aug, Dreamer-DANN, Dreamer-Std, and the high-speed semantic-policy variants. The success gain is therefore not explained by faster driving. It is more consistent with improved closed-loop stability and route completion under structural town shift.

The degenerate behavior of Sem-Adv-Aux is also visible in the speed and lane metrics. Its mean speed is close to zero in both towns, while its lane-invasion rate is high. This confirms that semantic-feature adversarial regularization alone does not produce a useful driving representation under the tested configuration.

\subsection{Termination-Mode Analysis}

\begin{figure*}[!t]
    \centering
    \captionsetup{font=footnotesize,skip=2pt}
    \includegraphics[width=\textwidth]{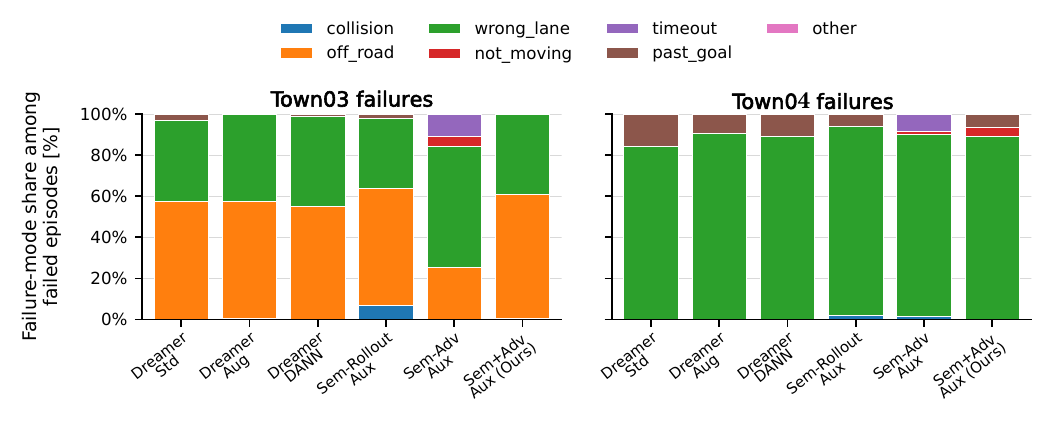}
    \caption{Failure-mode composition on held-out Town03 and Town04,
    pooled across the five training seeds and normalized among failed
    episodes only. Each bar reports the relative share of non-goal
    termination modes for a given method and town, conditional on the
    episode not reaching the goal. This figure describes how methods fail,
    not how often they fail, and should therefore be interpreted together
    with the success rates in \autoref{tab:town3_compact} and~\ref{tab:town4_compact}, since methods contribute different numbers of failed episodes.}
    \label{fig:failure_modes}
\end{figure*}

\autoref{fig:failure_modes} reports the composition of non-goal termination modes on
Town03 and Town04. The bars are normalized among failed episodes only, so the figure should be read as a conditional failure-mode diagnostic rather than as an overall performance metric. Overall task-completion performance is reported in \autoref{tab:town3_compact} and \ref{tab:town4_compact}.

On Town03, several low-success methods have low collision rates, but this does not imply safe or competent driving. These agents often terminate through off-road, wrong-lane, stuck, timeout, or other non-goal outcomes. Dreamer-Std is a representative example: it records a near-zero collision rate but has low success and short traveled distance, indicating poor coverage rather than higher route-completion success in this protocol.

The semantic-policy variants fail differently. They often travel more actively than the nearly stationary variants, but their low success and elevated adverse termination rates suggest unsafe or unstable control. This supports the design choice of keeping the actor and critic on $f_t$ while using $s_t$ only for auxiliary regularization.

On Town04, \autoref{tab:town3_compact} and \ref{tab:town4_compact} show that Sem+Adv-Aux, Dreamer-DANN, and Dreamer-Aug achieve high task-completion rates, with Sem+Adv-Aux obtaining the highest mean success. \autoref{fig:failure_modes} complements these success-rate results by showing the residual failure modes among episodes that do not reach the goal. Thus, the termination-mode analysis should not be read as a standalone ranking; it supports the main result by clarifying how failures are distributed after conditioning on failure.

\subsection{Route-Level Diagnostics}

\begin{table}[t]
\centering
\caption{Per-route held-out success rate. Each entry reports successful episodes divided by the number of evaluation episodes assigned to that route, aggregated across the five training seeds. Route IDs R0–R5 correspond to the six fixed held-out routes in each target town.}
\label{tab:per_route_success}
\scriptsize
\setlength{\tabcolsep}{2.5pt}
\renewcommand{\arraystretch}{1.05}
\begin{tabular}{llcccccc}
\hline
\textbf{Method} & \textbf{Town} & \textbf{R0} & \textbf{R1} & \textbf{R2} & \textbf{R3} & \textbf{R4} & \textbf{R5} \\
\hline
Sem+Adv-Aux (Ours)        & Town03 & 69 & 55 & 61 & 25 & 27 & 23 \\
Dreamer-Aug               & Town03 & 98 & 74 & 0  & 0  & 0  & 0  \\
Dreamer-DANN              & Town03 & 88 & 17 & 55 & 0  & 5  & 0  \\
Dreamer-Std               & Town03 & 12 & 0  & 0  & 0  & 17 & 1  \\
Sem-Rollout-Aux           & Town03 & 64 & 0  & 2  & 0  & 7  & 0  \\
Sem-Adv-Aux               & Town03 & 0  & 0  & 0  & 0  & 0  & 0  \\
SemSty-NoLoss             & Town03 & 34 & 0  & 0  & 0  & 23 & 5  \\
Full-Aux                  & Town03 & 26 & 19 & 0  & 0  & 3  & 0  \\
Policy-Sem-NoLoss         & Town03 & 1  & 0  & 0  & 0  & 0  & 0  \\
Small-GRL Semantic Policy & Town03 & 2  & 0  & 0  & 0  & 2  & 15 \\
Large-GRL Semantic Policy & Town03 & 1  & 0  & 0  & 0  & 3  & 0  \\
\hline
Sem+Adv-Aux (Ours)        & Town04 & 40 & 98 & 89 & 90  & 100 & 98  \\
Dreamer-Aug               & Town04 & 96 & 96 & 0  & 92  & 100 & 98  \\
Dreamer-DANN              & Town04 & 64 & 67 & 75 & 98  & 95  & 100 \\
Dreamer-Std               & Town04 & 0  & 0  & 0  & 4   & 100 & 100 \\
Sem-Rollout-Aux           & Town04 & 0  & 0  & 33 & 100 & 98  & 98  \\
Sem-Adv-Aux               & Town04 & 0  & 0  & 0  & 0   & 0   & 0   \\
SemSty-NoLoss             & Town04 & 56 & 2  & 5  & 100 & 82  & 90  \\
Full-Aux                  & Town04 & 0  & 51 & 0  & 100 & 100 & 98  \\
Policy-Sem-NoLoss         & Town04 & 0  & 0  & 0  & 38  & 100 & 100 \\
Small-GRL Semantic Policy & Town04 & 9  & 0  & 0  & 96  & 100 & 98  \\
Large-GRL Semantic Policy & Town04 & 5  & 33 & 0  & 92  & 100 & 100 \\
\hline
\end{tabular}
\end{table}

\autoref{tab:per_route_success} reports route-level diagnostics. These route-level diagnostics aggregate all evaluation episodes across the five training seeds. They are descriptive because each route has fewer episodes than the town-level aggregate, but they show whether the aggregate gain is distributed across routes or concentrated on a small subset.

On Town03, the proposed method improves aggregate success by increasing coverage on several difficult routes, but it does not solve every route. In the current heatmap, Sem+Adv-Aux performs well on several route IDs. The method improves zero-shot transfer but does not eliminate route-specific failure modes in the harder held-out town.

On Town04, Sem+Adv-Aux shows broader route coverage. It performs strongly on most visualized routes and avoids the severe route-specific collapse visible for Dreamer-Aug on one route. However, Dreamer-Aug and Dreamer-DANN remain competitive on other routes. The route-level results therefore support a nuanced interpretation: the proposed method improves aggregate transfer by broadening route coverage, not by uniformly dominating every route. The heatmaps show route IDs 0-5 because the evaluation protocol uses six fixed held-out routes per target town.

\subsection{Source-Training Diagnostics}

\begin{figure*}[!t]
    \centering
    \captionsetup{font=footnotesize,skip=2pt}
    \includegraphics[width=\textwidth]{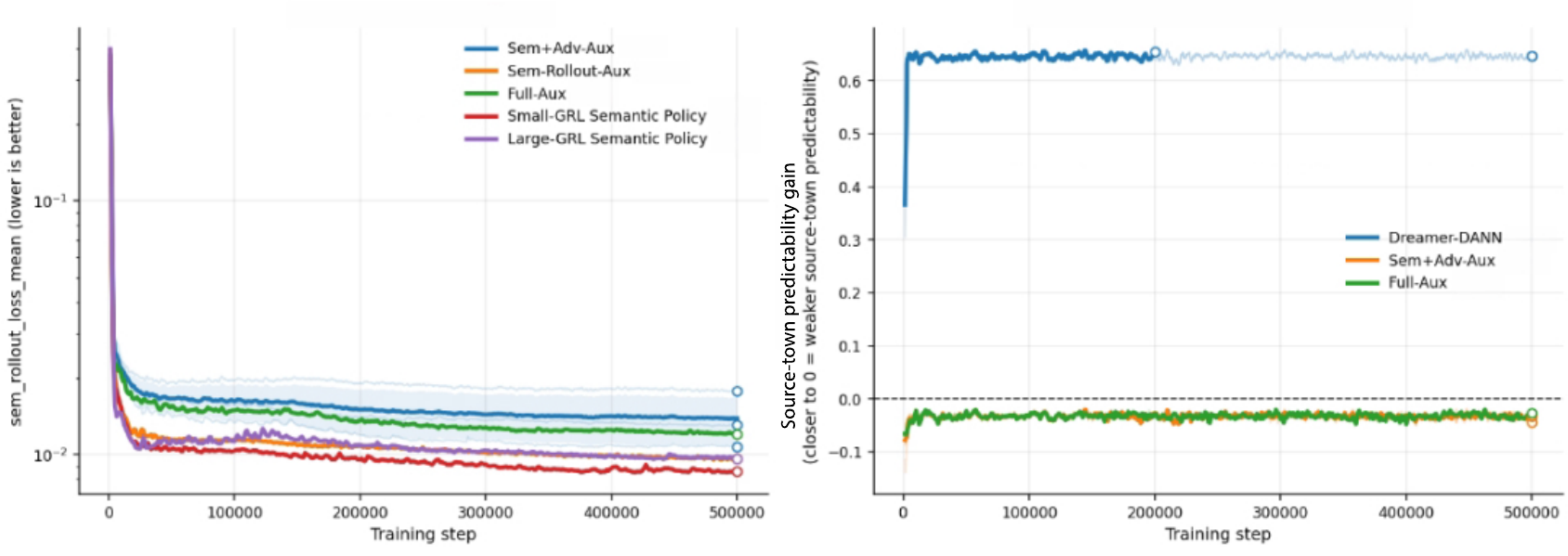}
    \caption{Right, source-town predictability diagnostic for the feature used by the domain head. The plotted value is a domain-predictability gain measure, where values near 0 indicate weak source-town predictability. Larger positive values indicate that the source-town label remains more predictable from the supervised feature. This source-only diagnostic is a training sanity check and is not evidence of invariance to Town03 or Town04.}
    \label{fig:train_diagnostics_appendix}
\end{figure*}

\autoref{fig:train_diagnostics_appendix} reports source-training diagnostics for the auxiliary losses. The semantic rollout loss decreases for rollout-enabled variants, confirming that the OpenCLIP-prediction auxiliary is being optimized during source-town training. The source-town predictability diagnostic shows how much town information remains available to the domain head attached to each feature.

These diagnostics are useful sanity checks, but they are not direct evidence of held-out-town invariance. The source-town labels are only $\{\mathrm{Town05}, \mathrm{Town06}\}$, so a low source-town predictability value means that the diagnostic head cannot distinguish the two source towns from the supervised feature. It does not prove that Town03 or Town04 are represented invariantly. The held-out-town claim of the paper should therefore rest on the closed-loop evaluation results in \autoref{tab:town3_compact} and~\ref{tab:town4_compact}, \autoref{fig:success_ft_vs_st}, and the seed-paired comparison in \autoref{tab:paired_success}.

\subsection{Summary of Findings}

Across the matched Dreamer-family variants, the main finding is that semantic rollout supervision and semantic-feature town-adversarial regularization are complementary. Semantic rollout alone improves over the plain Dreamer baseline but remains far below the proposed method. Semantic-feature town-adversarial regularization alone collapses. The combined Sem+Adv-Aux model achieves the highest mean held-out success on both Town03 and Town04 while retaining the standard Dreamer control feature for actor-critic learning.

The results also show that the semantic branch should not be used naively as the policy input. Variants that feed $s_t$ directly to the actor and critic perform poorly, especially on Town03. The best-performing configuration uses the semantic and context branches as training-time regularizers and keeps control on the standard Dreamer feature $f_t$.

Overall, the evidence supports a conclusion that  semantic future prediction combined with semantic-branch town-adversarial regularization improves zero-shot cross-town task completion within a matched Dreamer-style world-model family.
\section{Threats to Validity and Limitations} \label{sec:limitations}
This study is designed for controlled mechanism isolation rather than leaderboard-level CARLA driving. Four limitations are central to interpreting the results.

\textbf{Environmental scope.} The primary evaluation isolates cross-town structural variation under fixed ClearNoon weather and ego-only scenes. It does not include dynamic traffic, pedestrians, sensor corruption, weather variation, or real-world deployment. The results therefore support claims about the evaluated CARLA protocol, not general autonomous-driving robustness.

\textbf{Navigation interface.} The policy does not receive route commands, waypoints, target vectors, or privileged map inputs. The environment uses an internal reference path for reward, termination, and success. This creates a fixed-route closed-loop completion task and can be partially observable at intersections where multiple maneuvers are visually plausible.

\textbf{Domain-generalization scope.} The source-domain adversarial label is the binary Town05/Town06 identity. Reducing predictability of this label does not prove invariance to arbitrary unseen towns. Held-out generalization is assessed empirically through closed-loop evaluation in Town03 and Town04.

\textbf{Comparison scope.} The comparison is restricted to matched Dreamer-family variants. This design isolates the proposed auxiliary losses under a shared learner, reward, action space, and evaluation protocol, but it does not establish competitiveness with CARLA leaderboard systems, imitation-learning pipelines, route-conditioned planners, privileged-teacher methods, or multi-sensor driving stacks.

\section{Conclusion}

This paper studied controlled zero-shot cross-town transfer for a Dreamer-style world-model driving agent in CARLA. Agents were trained in Town05 and Town06 and evaluated directly in held-out Town03 and Town04 under a fixed camera/action interface, fixed ClearNoon weather, ego-only scenes, and matched route pools.

The proposed method combines two training-time auxiliary objectives: multi-horizon prediction of frozen OpenCLIP image embeddings along imagined latent rollouts and source-town adversarial regularization on a semantic projection of the recurrent latent state. The actor and critic retain the standard Dreamer control feature, so the auxiliary branches regularize representation learning without changing the deployed policy interface.

Under the evaluated protocol, the combined auxiliary design improves held-out task-completion success over matched Dreamer-style baselines and outperforms semantic rollout alone, source-town adversarial regularization alone, and direct semantic-policy control. The results support semantic rollout prediction and source-town adversarial regularization as useful mechanisms for improving transfer inside this controlled world-model setting. Future work should test the approach with route-conditioned navigation, additional towns, dynamic traffic, pedestrians, weather variation, stronger non-Dreamer baselines, and real-world or higher-fidelity closed-loop evaluation.

\bibliographystyle{IEEEtran}
\bibliography{sem_sty_paper}

\begin{IEEEbiography}
[{\includegraphics[width=1in,height=1.25in,clip,keepaspectratio]{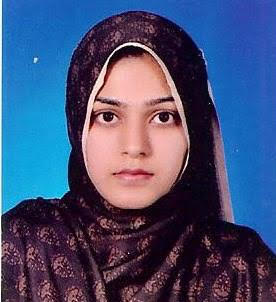}}]{Feeza Khan Khanzada} (Student Member, IEEE) received the B.E. degree in computer systems engineering from Mehran University of Engineering and Technology, Pakistan, and the M.S. degree in computer and information engineering. She is currently a graduate student research assistant and Ph.D. candidate at the University of Michigan–Dearborn. Her research interests include robotics, autonomous vehicles, deep learning for perception and control, probabilistic modeling, reinforcement learning, and robust decision-making in complex environments. She previously held research positions at Freie Universität Berlin and the University of Bath, working on machine learning, computer vision, and intelligent systems. Prior to her academic research roles, she was a software programmer with Fateh Motors Ltd., where she contributed to software development and system integration.
\end{IEEEbiography}

\begin{IEEEbiography}
[{\includegraphics[width=1in,height=1.25in,clip,keepaspectratio]{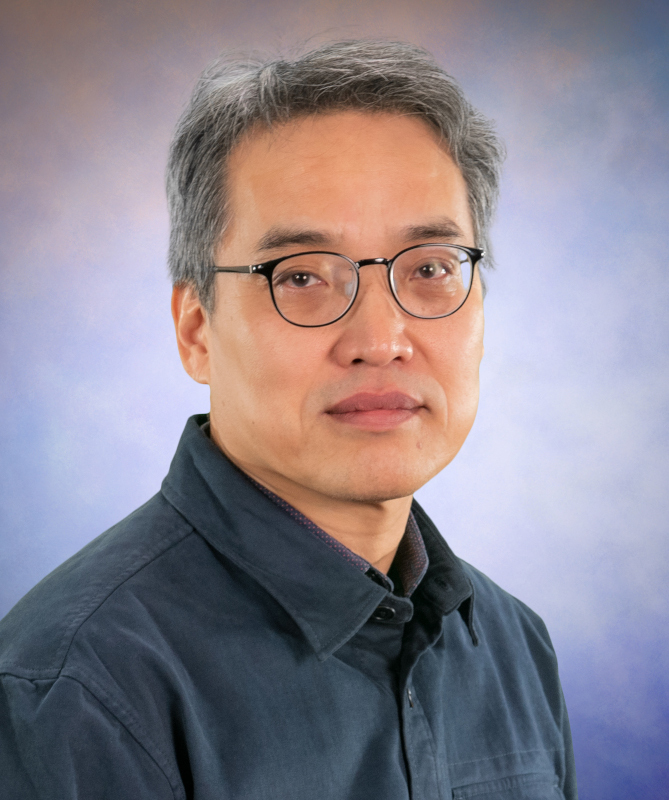}}]{Jaerock Kwon}
(Senior Member, IEEE) received the B.S. and M.S. degrees in Electronic Communication Engineering from Hanyang University, Seoul, South Korea, in 1992 and 1994, respectively, and the Ph.D. degree in Computer Engineering from Texas A\&M University, College Station, USA, in 2009. From 1994 to 2004, he worked at LG Electronics, SK Teletech, and Qualcomm Internet Services. From 2009 to 2010, he was a Professor at the Department of Electrical and Computer Engineering, Kettering University, Flint, MI, USA. Since 2010, he has been a Professor at the Department of Electrical and Computer Engineering, University of Michigan–Dearborn, MI, USA. His research interests include mobile robotics, autonomous vehicles, and artificial intelligence. His awards and honors include the Outstanding Researcher Award, the Faculty Research Fellowship (Kettering University), and the SK Excellent Employee (SK Teletech). He served as the President for the Korean Computer Scientists and Engineers Association in America (KOCSEA) in 2020, 2021, and 2025.
\end{IEEEbiography}

\vfill

\end{document}